\documentclass[runningheads]{llncs}
\usepackage[T1]{fontenc}
\usepackage{amsmath} 
\usepackage{amssymb}
\usepackage{graphicx}
\usepackage{cite}
\usepackage{hyperref}
\usepackage{bm}
\usepackage{multirow}
\begin{document}
\title{Enhancing Foundation Models for Imbalanced SAR Ship Classification via Targeted Oversampling}
%
%
\author{Ch Muhammad Awais\inst{1}\orcidID{0009-0001-4589-4103} \and
Marco Reggiannini\inst{1}\orcidID{0000-0002-4872-9541} \and
Davide Moroni\inst{1}\orcidID{0000-0002-5175-5126}}
\authorrunning{C. Awais et al.}
%
\institute{Institute of Science and Technology, National Research Council, Pisa, Italy
\email{fname.lname@cnr.it}}
\maketitle              
\begin{abstract}
Remote-sensing foundation models offer strong representations for SAR imagery, but their behavior under severe long-tail class imbalance is still not well characterized. We benchmark DOFA and SAR-JEPA on the imbalanced OpenSARShip dataset and compare them with ImageNet-pretrained baselines under a fixed, training-efficient protocol that keeps the backbone frozen. To mitigate imbalance without fine-tuning, we apply four oversampling methods in embedding space exclusively to minority classes and train a lightweight classifier head on the augmented embeddings. Across both foundation models, oversampling improves Macro-F1 and test accuracy relative to their respective baselines, with the largest Macro-F1 gains observed for DOFA using ADASYN (34.39 to 38.56) and for SAR-JEPA using SVM-SMOTE (25.89 to 32.30). We also report class-wise behavior, showing that aggregate improvements can coexist with persistent failures on specific rare classes. Code for embedding extraction and reproducible multi-seed evaluation is provided to support rapid experimentation on free-tier hardware.
\end{abstract}
\section{Introduction}
Synthetic Aperture Radar (SAR) enables all-weather maritime monitoring, making it suitable for ship classification in operational settings. However, practical SAR ship datasets are typically long-tailed, with a few dominating vessel categories, while rare types remain underrepresented \cite{awais2025survey}. Under this imbalance, high test accuracy can coexist with poor recognition of minority classes \cite{awais2024deep}. SAR target recognition has been studied with both traditional machine learning approaches \cite{ikeuchi1996invariant, fu2018aircraft, meth1999feature, nicoli2008shape, park2012new} and modern deep learning methods \cite{8127014, s19010063, rs13112091, 9782459, 10433638, 10642068, Awais_2025_WACV}, yet severe imbalance remains a persistent practical obstacle.

Deep learning models generally benefit from large and diverse labeled datasets, but collecting SAR ship imagery at scale is difficult and costly. As a result, prior work often relies on strategies such as transfer learning \cite{s19010063, RELEKAR20214594}, data augmentation \cite{rs16071299, 9784428}, multimodal fusion \cite{10457854, 10440349, 10189816}, and domain adaptation \cite{10246308, awais2025sar}. More recently, foundation models pretrained on remote sensing data have emerged as a promising alternative to ImageNet-pretrained backbones for Earth observation tasks \cite{xiao2025foundation, 10916803}. Despite this promise, there is limited evidence on how such representations behave on SAR ship classification when class distributions are highly skewed, and whether their benefits persist when evaluation emphasizes minority performance.

A natural way to address imbalance without collecting new labels is to rebalance the training distribution. Oversampling methods such as SMOTE \cite{chawla2002smote, awais2025feature} generate synthetic minority samples by interpolating between nearest neighbors in feature space, avoiding simple duplication. Several variants refine the synthesis process. SVM-SMOTE targets regions near decision boundaries, KMeans-SMOTE synthesizes within clusters, SMOTE-ENN removes ambiguous samples after oversampling, and ADASYN allocates more synthesis to locally difficult regions \cite{kovacs2019smote}. While these methods are well studied for conventional feature spaces, their behavior on embeddings produced by large pretrained models has received comparatively little attention, particularly for SAR where minority classes can be one to two orders of magnitude smaller than the majority ones.

This work addresses two questions. First, do remote-sensing foundation model embeddings provide better balanced performance than ImageNet-pretrained baselines under severe class imbalance? Second, can minority-only oversampling in embedding space artificially solve the balance issue in the original dataset and generate a reliable classifier, with high performance across categories without end-to-end backbone fine-tuning? To answer these questions, we benchmark Dynamic One-For-All (DOFA) \cite{xiong2024neural} and SAR-JEPA \cite{li2024predicting} on OpenSARShip \cite{li2017opensarship} under a controlled protocol. Embeddings are extracted once from a frozen backbone, oversampling is applied only to minority classes in feature space, and a lightweight classifier head is trained on the augmented embeddings. This design supports rapid experimentation and reproducible evaluation on free-tier hardware. In summary, our contributions are:
\begin{enumerate}
    \item A benchmark of remote-sensing foundation model embeddings on an imbalanced SAR ship classification task, reporting Macro-F1 alongside accuracy.
    \item A training-efficient imbalance mitigation pipeline that performs minority-only oversampling in embedding space and improves balanced performance without backbone fine-tuning.
    \item An open-source, Colab-friendly implementation that pre-extracts embeddings and supports reproducible multi-seed evaluation. \hyperlink{https://github.com/cm-awais/SARShipfoundationModels}{https://github.com/cm-awais/SARShipfoundationModels}
\end{enumerate}

\section{Methodology}
\begin{figure}[htbp]
    \centering
    \includegraphics[width=1\linewidth]{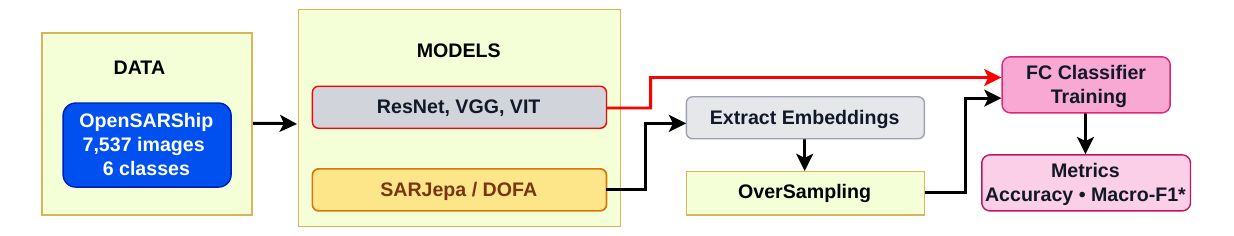}
    \caption{Overview of the proposed evaluation pipeline. Given OpenSARShip, we extract frozen embeddings from each backbone (ImageNet baselines and Earth Observation foundation models), apply minority-only oversampling in embedding space, and train a lightweight fully connected classifier head on the augmented embeddings.}
    \label{fig:methodology}
\end{figure}

Fig.~\ref{fig:methodology} summarises the pipeline. The backbone is kept frozen throughout: embeddings are extracted once, minority-only oversampling is applied in feature space, and a lightweight classifier head is trained on the augmented embeddings. This design avoids costly backbone fine-tuning and keeps experimentation tractable on free-tier hardware.

\subsection{Dataset and Foundation Models}
The proposed method is evaluated on OpenSARShip \cite{li2017opensarship}, a SAR ship classification dataset with six classes: Cargo (5303 samples), Tanker (1825), Dredging (142), Fishing (139), Passenger (66), and Tug (62). The dataset is split into train, validation, and test sets using an 80/10/10 ratio. Given a pretrained encoder $f(\cdot)$, each SAR chip is mapped to a $d$-dimensional embedding, which is extracted for all splits and saved to disk before any oversampling or classifier training takes place.

We benchmark two remote-sensing foundation models: Dynamic One-For-All (DOFA) \cite{xiong2024neural}, a unified multimodal framework designed for diverse earth observation tasks, and SAR-JEPA \cite{li2024predicting}, a joint-embedding predictive architecture trained directly on SAR data. Both are compared against three ImageNet-pretrained baselines, ResNet, VGG \cite{mascarenhas2021comparison}, and ViT-224 \cite{yuan2021tokens}, evaluated under the same protocol.

\subsection{Oversampling in Embedding Space}
Classes with less than 200 training samples are treated as minority classes. The subset related to the minority classes is thus given by $\mathcal{M} =$ Dredging, Fishing, Passenger, Tug. For each minority class, oversampling is applied exclusively within that class in feature space, until the class size reaches 3 times its original value. Majority classes are never modified, and validation and test embeddings are left untouched.

All four oversampling methods generate synthetic points by interpolating between same-class nearest neighbours in $\mathbb{R}^{d}$:
\begin{equation}
\tilde{\mathbf{z}} = \mathbf{z}_i + \lambda(\mathbf{z}_j - \mathbf{z}_i), \qquad \lambda \sim \mathcal{U}(0,1).
\end{equation}
The methods differ in how the pairs $(\mathbf{z}_i, \mathbf{z}_j)$ are selected: SVM-SMOTE restricts synthesis to regions near the support-vector decision boundary; KMeans-SMOTE first clusters the minority classes, then synthesises within each cluster, keeping synthetic samples close to the true data manifold; SMOTE-ENN starts with a KMeans-SMOTE approach, then it additionally removes ambiguous samples using Edited Nearest Neighbours; and ADASYN weights synthesis density by local class difficulty i.e., generating more samples in regions where minority examples overlap with the majority ones \cite{kovacs2019smote}.


\subsection{Classifier and Training Protocol}
The classifier head $g(\cdot)$ is a small three-layer perceptron with batch normalisation, ReLU activations, and dropout ($p=0.1$). It is trained on the augmented training set using cross-entropy loss with Adam (learning rate $10^{-3}$, batch size 256, 50 epochs). Results are reported as averages over three random seeds, varying model initialisation and sampler randomness while keeping the data split fixed.
\begin{table}[htbp]
\centering
\caption{Overall performance metrics.}
\label{tab:overall}
\setlength{\tabcolsep}{6pt}
\renewcommand{\arraystretch}{1.0}
\begin{tabular}{llcc}
\hline
\textbf{Model} & \textbf{Method} & \textbf{Accuracy} & \textbf{Macro-F1} \\
\hline
DOFA     & Base   & 60.65$\pm$1.12 & 34.39$\pm$0.26 \\
DOFA     & ADASYN & 72.42$\pm$0.84 & \textbf{38.56$\pm$1.85} \\
DOFA     & KMeans & \textbf{72.82$\pm$0.75} & 35.59$\pm$0.62 \\
DOFA     & SENN   & 66.96$\pm$1.17 & 33.85$\pm$1.23 \\
DOFA     & SVM    & 72.32$\pm$1.36 & 36.87$\pm$1.05 \\
\hline
SAR-JEPA & Base   & 48.02$\pm$2.53 & 25.89$\pm$1.20 \\
SAR-JEPA & ADASYN & 69.74$\pm$0.98 & 29.28$\pm$0.54 \\
SAR-JEPA & KMeans & \textbf{71.76$\pm$1.22} & 30.86$\pm$2.64 \\
SAR-JEPA & SENN   & 66.37$\pm$0.14 & 28.11$\pm$1.25 \\
SAR-JEPA & SVM    & 71.20$\pm$0.42 & \textbf{32.30$\pm$0.26} \\
\hline
ResNet-50 & Base  & 70.24$\pm$0.25 & 13.75$\pm$0.42 \\
VGG-16    & Base  & 71.30$\pm$0.33 & 19.93$\pm$0.11 \\
ViT-224   & Base  & 73.74$\pm$0.39 & \textbf{27.03$\pm$0.01} \\
\hline
\multicolumn{4}{p{6.5cm}}{\footnotesize Base: baseline; KMeans: KMeans-SMOTE; SENN: SMOTEENN; SVM: SVMSMOTE} \\
\hline
\end{tabular}
\end{table}
\section{Results}
We evaluated five classification pipelines across two foundation models and three ImageNet baselines. Table~\ref{tab:overall} summarizes overall test accuracy and Macro-F1 scores. Among baseline architectures, ViT-224 achieved the highest accuracy at 73.74\% with a Macro-F1 of 27.03, substantially outperforming ResNet and VGG.

The DOFA baseline achieved lower accuracy than the ImageNet baselines in Table~\ref{tab:overall}, but oversampling improved it substantially. KMeans-SMOTE produced the best accuracy at 72.82\%, while ADASYN achieved the highest Macro-F1 score of 38.56. Both ADASYN and SVMSMOTE converged to similar accuracy levels around 72.3\%. SMOTEENN showed degraded performance relative to other oversampling variants, reaching only 66.96\% accuracy.

SAR-JEPA baseline performance was markedly weaker, achieving 48.02\% accuracy and 25.89 Macro-F1. Oversampling substantially improved results, with KMeans-SMOTE reaching 71.76\% accuracy and SVMSMOTE attaining 32.30 Macro-F1. The large gap between SAR-JEPA baseline and oversampled variants suggests SAR-JEPA is more sensitive to the choice of the oversampling method under our fixed classifier setup. ADASYN and SVMSMOTE both reached approximately 71\% accuracy, while SMOTEENN again underperformed.

\begin{table}[ht]
\centering
\caption{Per-class F1-scores across configurations.}
\label{tab:perclass}
\setlength{\tabcolsep}{4pt}
\renewcommand{\arraystretch}{1.0}
\begin{tabular}{lccccc}
\hline
\textbf{Class} & \textbf{Baseline} & \textbf{ADASYN} & \textbf{KMeans} & \textbf{SENN} & \textbf{SVM} \\
\hline
\multicolumn{6}{p{6.5cm}}{\textbf{DOFA}} \\
Cargo     & 73.15$\pm$1.27 & 82.52$\pm$0.67 & 83.09$\pm$0.64 & 81.32$\pm$0.34 & \textbf{83.54$\pm$0.19} \\
Dredging  & 18.74$\pm$3.64 & \textbf{19.93$\pm$5.09} &  8.28$\pm$4.46 & 19.71$\pm$1.14 &  8.99$\pm$0.14 \\
Fishing   & 27.47$\pm$2.05 & 24.70$\pm$5.47 & 30.56$\pm$2.82 & 32.03$\pm$6.72 & \textbf{34.31$\pm$4.52} \\
Passenger & 24.98$\pm$7.47 & \textbf{38.41$\pm$7.17} & 28.83$\pm$3.88 & 26.08$\pm$2.40 & 30.79$\pm$1.12 \\
Tanker    & \textbf{52.40$\pm$1.39} & 51.30$\pm$0.75 & 49.91$\pm$1.64 & 30.75$\pm$2.51 & 49.22$\pm$0.63 \\
Tug       & 19.00$\pm$3.24 & 18.53$\pm$3.57 & 18.93$\pm$10.1 & 20.66$\pm$7.04 & \textbf{24.50$\pm$0.71} \\
\hline
\multicolumn{6}{p{6.5cm}}{\textbf{SAR-JEPA}} \\
Cargo   & 63.57$\pm$2.37 & 81.94$\pm$0.51 & \textbf{82.83$\pm$0.56} & 81.34$\pm$0.26 & 82.46$\pm$0.16 \\
Dredging  & 11.24$\pm$2.18 &  8.25$\pm$2.24 &  5.08$\pm$0.90 & \textbf{18.29$\pm$2.99} & 13.50$\pm$3.09 \\
Fishing   & 22.90$\pm$2.84 & 30.45$\pm$0.94 & 36.02$\pm$8.25 & 27.44$\pm$3.45 & \textbf{39.92$\pm$1.27} \\
Passenger &  7.04$\pm$1.41 &  6.90$\pm$4.88 &  6.70$\pm$1.83 & \textbf{10.82$\pm$1.84} &  0.00$\pm$0.00 \\
Tanker    & \textbf{44.57$\pm$1.43} & 34.50$\pm$0.92 & 38.84$\pm$0.43 & 25.65$\pm$2.26 & 40.08$\pm$2.10 \\
Tug       &  9.61$\pm$2.55 & 17.39$\pm$0.00 & 19.05$\pm$0.00 & 11.21$\pm$3.00 & \textbf{21.64$\pm$0.83} \\
\hline
\end{tabular}
\end{table}

Per-class results reveal the underlying challenges of the dataset. Table~\ref{tab:perclass} presents F1-scores across vessel types and methods, highlighting differences between backbones. The Cargo class demonstrates consistently strong performance across methods. For DOFA, Cargo F1 is above 81\% for all methods, while for SAR-JEPA it ranges from 63.57\% (Baseline) to 82.83\% (KMeans-SMOTE). This suggests that majority classes benefit uniformly from the oversampling strategies employed.

DOFA exhibits more balanced per-class performance overall. Passenger shows the largest spread across oversampling methods within each backbone. For DOFA, ADASYN yields the strongest Passenger F1 (38.41\%), whereas for SAR-JEPA some oversampling choices degrade Passenger severely, including a null F1 in case SVMSMOTE is adopted, despite a moderate overall accuracy. This highlights why overall accuracy can be misleading under severe imbalance and why per-class reporting is necessary even when Macro-F1 is included.

Minority classes follow distinct patterns by method. The Dredging class shows marked volatility in DOFA, dropping to 8.28\% with KMeans-SMOTE but reaching 19.93\% with ADASYN. SAR-JEPA performs even worse on Dredging, with all methods producing F1-scores below 18.29\%. The Fishing class exhibits more consistent improvements, with SAR-JEPA SVMSMOTE achieving the highest performance at 39.92\%. Tanker class performance degrades with SMOTEENN for both architectures, suggesting this method may not suit denser minority manifolds.

SAR-JEPA baseline underperforms substantially across most classes, particularly on Passenger (7.04\%) and Tug (9.61\%), indicating that this backbone is more sensitive to class imbalance under our fixed classifier setup. DOFA baseline provides more competitive baseline scores, with Tanker at 52.40\% and Cargo at 73.15\%. The oversampling techniques improve SAR-JEPA performance across the board, but introduce high variance in specific class-method pairs. ADASYN and KMeans-SMOTE provide the most consistent improvements, though their effectiveness varies by class and architecture.

\section{Discussion}
This study benchmarks remote-sensing foundation model embeddings for imbalanced SAR ship classification and shows that applying proper minority-only oversampling methods in embedding space, can improve Macro-F1 in this benchmark, without backbone fine-tuning. Under a fixed classifier head and training protocol, oversampling consistently increases Macro-F1 for both DOFA and SAR-JEPA, while keeping the overall pipeline lightweight and reproducible.

A first outcome is that accuracy alone is not representative of performance under the OpenSARShip long-tail distribution. The ImageNet baselines obtain the highest accuracies, with ViT-224 reaching 73.74\%, yet their Macro-F1 scores remain low (Table~\ref{tab:overall}). This indicates that correct predictions are concentrated on the frequent classes. In contrast, DOFA exhibits a higher Macro-F1 at baseline despite lower accuracy, suggesting that its embedding space preserves more separability for minority classes even before any oversampling is applied.

A second outcome is that the choice of oversampling method affects accuracy and Macro-F1 differently, and the best configuration depends on the target metric. For DOFA, KMeans-SMOTE yields the highest accuracy, whereas ADASYN achieves the highest Macro-F1 (Table~\ref{tab:overall}). For SAR-JEPA, KMeans-SMOTE again maximizes accuracy, while SVM-SMOTE attains the best Macro-F1. This separation reinforces the need to report balanced metrics when comparing models for long-tail SAR classification, since improvements in accuracy can be decoupled from improvements on rare classes.

The class-wise view clarifies where these gains originate. Table~\ref{tab:perclass} shows that Cargo remains consistently strong across methods, which is expected given its large support. The more informative differences appear among the minority classes. Passenger is a particularly challenging class: DOFA combined with ADASYN yields the strongest Passenger F1, while SAR-JEPA remains weak and can fail entirely for specific oversampling choices. Dredging also remains challenging for both backbones, and its response to oversampling is inconsistent. Fishing follows a different profile, with larger improvements in some SAR-JEPA settings. These patterns indicate that minority classes are not interchangeable and that a single oversampling strategy may not be uniformly beneficial.

Since the backbone is frozen, all improvements are attributable to changes in the effective training distribution seen by the classifier head. Interpolation-based synthesis is likely to help when same-class neighborhoods in embedding space are locally coherent, because synthetic points expand the decision region with samples that remain consistent with the class manifold. Conversely, when a minority class is fragmented, overlapping, or poorly represented, synthetic points can amplify noise and degrade class discrimination. The per-class failures visible in Table~\ref{tab:perclass} are consistent with this limitation and motivate reporting class-wise results alongside aggregate metrics.

From a practical standpoint, this pipeline uses frozen pretrained embeddings as a lightweight testbed for imbalance mitigation, letting us leverage strong representations while avoiding the cost and complexity of end-to-end fine-tuning. It also enables rapid comparison of oversampling techniques under controlled conditions. In this benchmark, ADASYN is the strongest choice for maximizing Macro-F1 with DOFA, while SVM-SMOTE is the strongest choice for SAR-JEPA. If accuracy is prioritized, KMeans-SMOTE is the best configuration for both foundation models (Table~\ref{tab:overall}). These selections should still be validated against per-class behavior when the operational requirements emphasize specific rare categories.

There are several limitations to consider. Results are averaged over three random seeds with a fixed dataset split, so the reported variability reflects initialization and sampler randomness but does not capture split sensitivity. Oversampling is applied with a single minority threshold and a fixed target multiplier, even though different classes may require different augmentation strengths. Finally, feature-space interpolation assumes that linear combinations of neighbors remain plausible within the representation space, which may not hold equally for all classes and backbones.

Future work can extend this benchmark by making the oversampling policy class-adaptive, selecting both method and oversampling ratio using validation performance per class. Additional studies can combine embedding-space oversampling with loss reweighting or focal objectives while keeping the backbone frozen. A further direction is to introduce parameter-efficient fine-tuning, such as adapters or LoRA \cite{hu2022lora, wang2024flora}, to test whether the observed minority gains persist when representation learning is allowed to adjust.

\section{Conclusion}
This work provides a benchmark of remote-sensing foundation model embeddings for imbalanced SAR ship classification and evaluates minority-only oversampling in embedding space as a simple alternative to end-to-end fine-tuning. Under a fixed classifier head and training setup, oversampling improves balanced performance for both DOFA and SAR-JEPA. DOFA achieves its best Macro-F1 with ADASYN (38.56), while SAR-JEPA achieves its best Macro-F1 with SVM-SMOTE (32.30). In terms of accuracy, KMeans-SMOTE yields the strongest results for both foundation models (72.82 for DOFA and 71.76 for SAR-JEPA), highlighting that the configuration that maximizes accuracy is not necessarily the one that maximizes Macro-F1.

Class-wise results further show that improvements are not uniform across categories. Cargo remains consistently strong, while rare classes such as Passenger and Dredging remain challenging and can be sensitive to the oversampling choice, including occasional severe degradations for specific combinations. These findings support reporting both aggregate and per-class metrics when evaluating imbalanced SAR recognition systems. Future work will address the development of ad-hoc techniques, including class-adaptive oversampling and parameter-efficient adaptation.


%
%
%
\bibliographystyle{splncs04}
\bibliography{ICPR_2026_LaTeX_Templates/bibliography}
%




\end{document}